\documentclass[sigconf]{acmart}
\usepackage[ruled,vlined]{algorithm2e}
\usepackage{multirow}
\usepackage{booktabs,caption}
\usepackage[flushleft]{threeparttable}
\AtBeginDocument{%
  \providecommand\BibTeX{{%
    \normalfont B\kern-0.5em{\scshape i\kern-0.25em b}\kern-0.8em\TeX}}}

\copyrightyear{2021}
\acmYear{2021}
\setcopyright{acmcopyright}\acmConference[CIKM '21]{Proceedings of the 30th ACM
International Conference on Information and Knowledge Management}{November
1--5, 2021}{Virtual Event, QLD, Australia}
\acmBooktitle{Proceedings of the 30th ACM International Conference on Information
and Knowledge Management (CIKM '21), November 1--5, 2021, Virtual Event, QLD,
Australia}
\acmPrice{15.00}
\acmDOI{10.1145/3459637.3482403}
\acmISBN{978-1-4503-8446-9/21/11}

\begin{document}

\title{Zero-shot Relation Classification from Side Information}

\author{Jiaying Gong}
\affiliation{%
  \institution{Virginia Tech}
  \city{Blacksburg}
  \country{U.S.}}
\email{gjiaying@vt.edu}

\author{Hoda Eldardiry}
\affiliation{%
  \institution{Virginia Tech}
  \city{Blacksburg}
  \country{U.S.}}
\email{hdardiry@vt.edu}

\fancyhead{}
\renewcommand{\shortauthors}{Trovato and Tobin, et al.}

\begin{abstract}
  We propose a zero-shot learning relation classification (ZSLRC) framework that improves on state-of-the-art by its ability to recognize novel relations that were not present in training data. The zero-shot learning approach mimics the way humans learn and recognize new concepts with no prior knowledge. To achieve this, ZSLRC uses advanced prototypical networks that are modified to utilize weighted side (auxiliary) information. ZSLRC's side information is built from keywords, hypernyms of name entities, and labels and their synonyms. ZSLRC also includes an automatic hypernym extraction framework that acquires hypernyms of various name entities directly from the web. ZSLRC improves on state-of-the-art few-shot learning relation classification methods that rely on labeled training data and is therefore applicable more widely even in real-world scenarios where some relations have no corresponding labeled examples for training. We present results using extensive experiments on two public datasets (NYT and FewRel) and show that ZSLRC significantly outperforms state-of-the-art methods on supervised learning, few-shot learning, and zero-shot learning tasks. Our experimental results also demonstrate the effectiveness and robustness of our proposed model.
\end{abstract}




\maketitle

\section{Introduction}
Relation classification aims to infer the relation between two name entities in a sentence.
Supervised learning methods for relation classification have been widely used to classify relations based on training labeled data.
Distant supervision or crowdsourcing have been used to collect more examples with labels and train the model for relation classification.
However, these methods are limited by the quantity (for supervised) and quality (for distant-supervised) of the training data because manually labeling the data is time-consuming and labor-intensive, and data labeled by distant-supervision is noisy.
To overcome the problem of insufficient high-quality data, few-shot learning has been designed to require only few labeled sentences for training.
A lot of research has been done on few-shot learning for computer vision~\citep{Lifchitz_2019_CVPR, Li_2020_CVPR, Ye_2020_CVPR}, and some work also includes few-shot learning methods for relation classification~\cite{han-etal-2018-fewrel, article1, few-shot2020}. However, these works still require a few instances for training, and they still do not work when no training instances are available.

\begin{figure}[htp] 
 \center{\includegraphics[height=3.5cm,width=8cm]{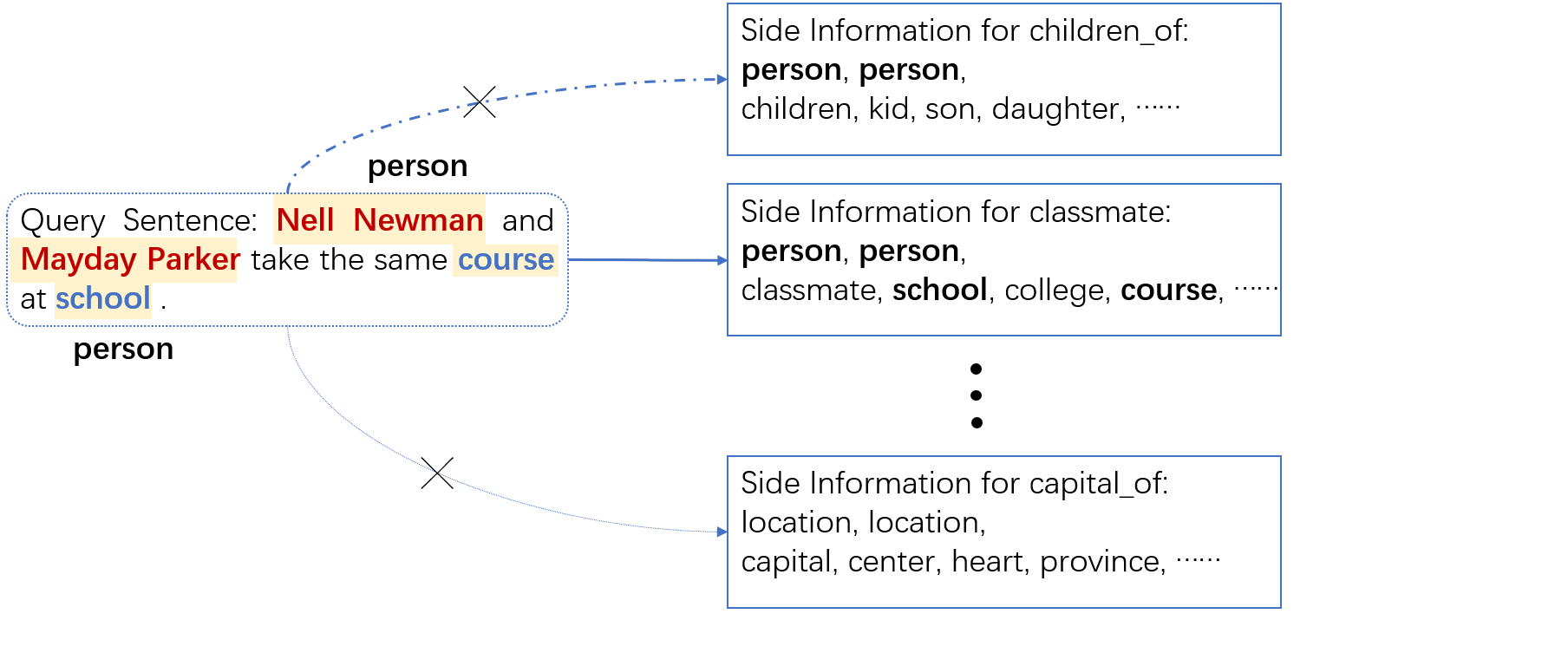}} 
 \caption{\label{fig:ZSLex} Example of relation classification based on side information.} 
 \end{figure}

Some work on open information extraction (OpenIE) discovers new relationships in open-domain corpora without labeling the data~\cite{angeli-etal-2015-leveraging}. OpenIE aims to extract relation phrases directly from the text. However, this technique can not effectively select meaningful relation patterns and discard irrelevant information.
Besides, this technique can not discover relations if the relation's name does not appear in the given sentence.
For example, OpenIE can not identify the relation of the sentence in Figure~\ref{fig:ZSLex}.

To address the limitations mentioned above, we focus on relation classification in the context of zero-shot learning.
Zero-shot learning (ZSL) is similar to the way humans learn and recognize new concepts.
It is a novel learning technique that does not use any exemplars of the unseen categories during training.
We propose a zero-shot learning model for relation classification (ZSLRC), which focuses on recognizing new relations with no corresponding labeled data available for training.
ZSLRC is modified on prototypical networks utilizing side (auxiliary) information. 
We construct weighted side information from labels and their synonyms, hypernyms of two name entities, and keywords from training sentences.
The ZSL-based model can recognize new relations based on the side information available for it instead of using a collection of labeled sentences.
We incorporate side information to enable our model to identify relations that never appear in the training datasets.
We also build an automatic hypernym extraction framework to help us acquire hypernyms of different entities directly from the web.
Details of side information construction are described in Section~\ref{section:SIA}.
 
Figure~\ref{fig:ZSLex} shows an example of how side information can be used for classifying relations.
Different side information is given for different relations.
The query sentence in the example has a relation of \textit{classmate\_of}, but the word classmate never appears in the sentence.
We first get the two name entities \textit{Nell Newman} and \textit{Mayday Parker} of the sentence and extract the hypernyms of the name entities \textit{person} and \textit{person} based on our proposed hypernym extraction module in Section~\ref{sec:ha}.
In this example, relation \textit{capital\_of} is eliminated because the hypernyms of \textit{capital\_of} should be \textit{location} and \textit{location}.
Then we extract the keywords \textit{course} and \textit{school} from the query sentence and compare the distance with the keywords in the side information box. 
In this way, relation \textit{children\_of} is eliminated.

To make relation classification effective in real-world scenarios, we design our model with the ability of classifying both relations with training instances and relations without any training instances. 
We modify the vanilla prototypical networks to deal with both scenarios and compare the distance between the query sentence and the weighted prototype.
If the exponential of the minus distance is above a threshold, we consider the query sentence has a new relation.
For new relations identification, we take the side information embedding from the query sentence and compare the distance of it with the side information embedding of new relations.
We conduct different experiments on both a noisy and a clean dataset and adding different percentages of new relations to evaluate the effectiveness and robustness of our proposed model.
Besides, we also evaluate our proposed model in supervised learning, few-shot learning, and zero-shot learning tasks. The results show that our proposed model outperforms other existing models in all three tasks.
The contributions of this paper can be summarized as follows:

\begin{itemize}
    \item We propose the first approach (ZSLRC) to enable zero-shot learning on relation classification without relying on other complex models that need to be learned and assumed to be 100\% accurate.
    
	\item ZSLRC uses side information including labels, keywords, and hypernyms of name entities, and it has been shown that our model can perform competitively using the weighted side information.
	We build an automatic hypernym extraction framework to extract hypernyms of words from the web.
	\item We modify prototypical networks to recognize new relations in addition to recognized previously known relations.
	Results show the effectiveness and robustness of our modified prototypical networks in different learning tasks.
	\item We demonstrate that our proposed model significantly outperforms state-of-the-art methods on supervised learning, few-shot learning, and zero-shot learning tasks. We ran extensive experiments on two datasets.
\end{itemize}


\section{Related Work}\label{LR}
\textbf{Supervised Relation Classification.}
Relation Classification aims to classify relations between entities.
Many existing relation classification methods are based on supervised learning, where neural networks are used to extract semantic features from text automatically. For example, convolutional neural networks (CNNs) are used to learn textual patterns~\cite{inproceedings, zeng-etal-2014-relation, nguyen-grishman-2015-relation, dos-santos-etal-2015-classifying, wang-etal-2016-relation, inproceedings1}.
Recurrent neural networks (RNNs) are used to better capture the sequential information present in the input data ~\cite{DBLP:journals/corr/ZhangW15a, zhou-etal-2016-attention, DBLP:journals/corr/NguyenG15}.
Graph neural networks (GNNs) are used to find dependencies and capture long-range relations between words ~\cite{zhang-etal-2018-graph, zhu-etal-2019-graph}.
Although these traditional Relation Classification methods have achieved promising results by taking advantage of supervised or distantly-supervised data, they exhibit a fundamental limitation since they all need large quantities of labeled training data.

\textbf{Open Relation Extraction.}
Many existing approaches focus on discovering new relationships in open-domain corpora. This is because traditional supervised RC can not find new relation types due to their limited ability to classify predefined relation types. Open RE or Open information extraction (OpenIE) aims to extract relation phrases directly from the text. For example, tagging-based methods~\cite{Jia2020HybridNT, cui-etal-2018-neural} and clustering-based methods~\cite{marcheggiani-titov-2016-discrete, wu-etal-2019-open} are used to discover new relation types.
Other work proposed Relational Siamese Networks to transfer relational knowledge from supervised OpenRE data to calculate similarity of unlabeled sentences for open relation clustering ~\cite{wu-etal-2019-open}. However, OpenRE can not effectively select meaningful relation patterns and discard irrelevant information. In the real world, methods that rely on predefined relation types are always known to lack of training data.

\textbf{Zero-shot Learning.} Zero-shot learning has been widely applied in computer vision~\cite{Yu_2020_CVPR, Keshari_2020_CVPR, Xie_2019_CVPR, Huynh_2020_CVPR, Rahman_2019_ICCV, Bustreo_2019_ICCV, Li2019RethinkingZL}.
Similar to zero-shot learning, few shot learning is well-studied in the field of relation classification~\cite{8258168, han-etal-2018-fewrel, article1, Gao2020NeuralSF, ye-ling-2019-multi,dong-etal-2020-meta}.
However, compared with zero-shot learning for computer vision and few-shot learning explored in relation classification, there exists little work towards zero-shot learning in the domain of natural language processing.
Some current work uses a transferable architecture to jointly represent and map event types in order to detect unseen event types~\cite{huang-etal-2018-zero}. Other work proposed a zero-shot learning method for relation extraction from webpages with unseen templates~\cite{lockard-etal-2020-zeroshotceres}.
However, this method solves a different problem, only predicting relation types in unseen structures of webpages instead of new relation types. 
The most related work to zero-shot learning for relation classification uses zero-shot learning to extract unseen relation types by listing questions that define the relation’s slot values~\cite{levy-etal-2017-zero}. However, this method requires external help, such as a question-answering dataset annotated by a human. In addition, this method assumes that (1) a good reading comprehension model is learned and that (2) all values extracted from this model are correct.
In contrast, our proposed model can identify new relation types without training sentences and does not need to rely on other models. We construct weighted side information to train the model without labeled training sentences. For example, some previous works use side information from knowledge graph or label to lower the noise and improve performance in distantly-supervised relation classification~\cite{vashishth-etal-2018-reside, hu-etal-2019-improving}.

\section{Methodology} \label{Meth}

\begin{figure*}[h] 
 \center{\includegraphics[height=9cm,width=\textwidth]{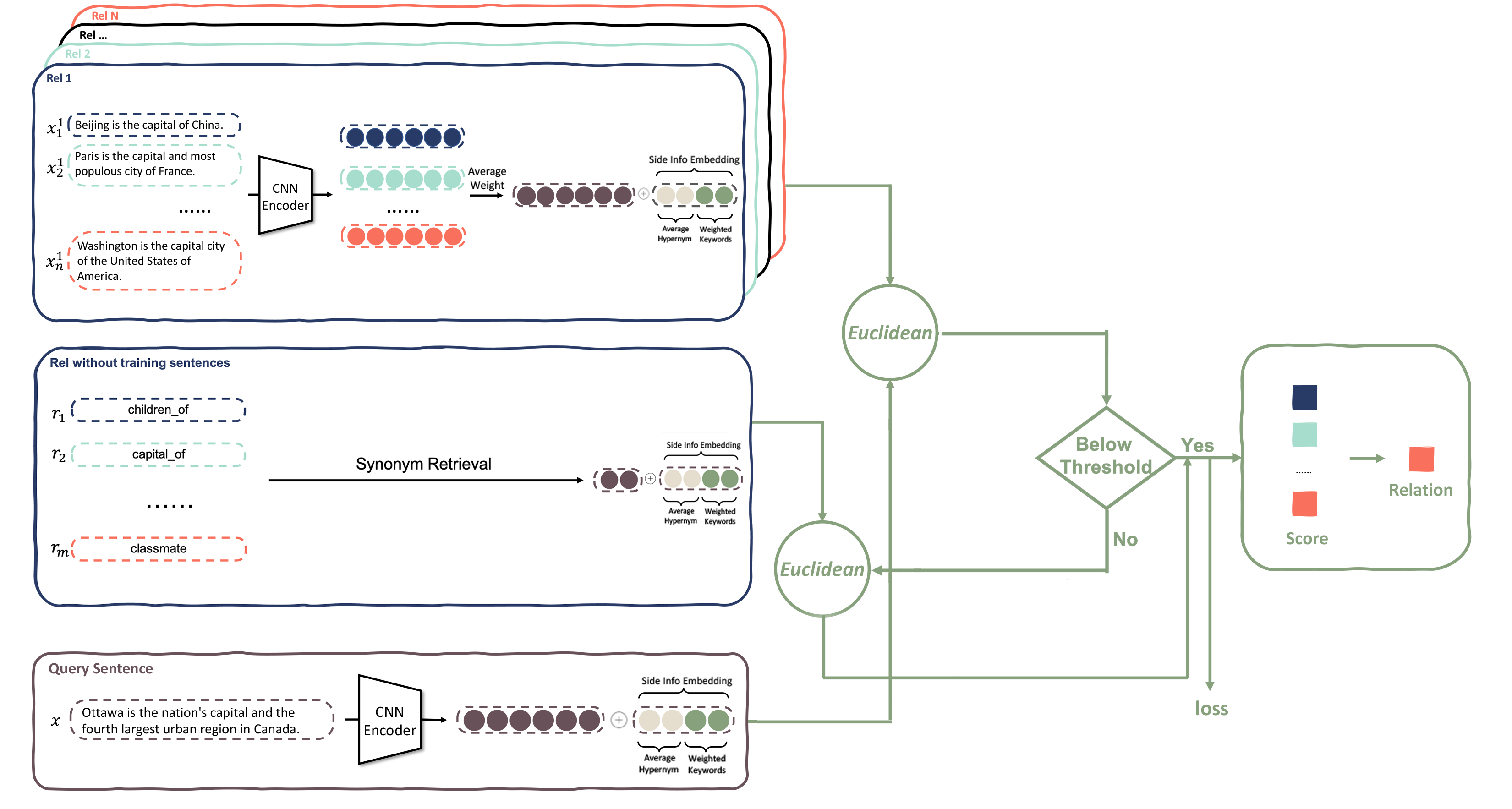}} 
 \caption{\label{fig:Model} Model of Zero-shot Learning for Relation Classification (ZSLRC)} 
 \end{figure*}

In this section, we introduce the overview of ZSLRC model. 
Figure~\ref{fig:Model} shows the architecture of zero-shot learning for relation classification. 
It consists of three parts: Sentence Encoder, Side Information Extraction and Prototypical Network with Weighted Side Information Embedding.
We describe these parts in detail below.

\subsection{Sentence Encoder}
The inputs of ZSLRC model are a set of sentences $\begin{Bmatrix}
x_{1},x_{2},x_{3},\cdots x_{n}
\end{Bmatrix}$ and its corresponding entity pair. 
For relations with training sentences, our model measures the probability of each relation $r'$ by measuring the distance between query sentences and the average weight of training sentence embeddings.
For relations without training sentences, the probability of $r'$ is done by measuring the distance between side information from query sentences and side information from relation types.
\subsubsection{\textbf{Word Embeddings}}
Word embeddings aim to map words or phrases from vocabulary to vectors of numerical forms.
The distributed representations are learned based on the usage of words, which allows words that are used in similar ways to result in having similar representations, naturally capturing syntactic and semantic meanings of the words.
In this paper, we first tokenize and lemmatize all words in a sentence, and a 50-dimension GloVe, a pre-trained global log-bilinear regression model for the unsupervised learning of word representations, is used as our initial word embeddings~\cite{pennington2014glove}.
If the words are out of vocabulary, they are randomly embedded first, and the vectors are updated while the model is training.
Word embedding vectors are updated through training the model.

\subsubsection{\textbf{Position Embeddings}}
Word positions also play an essential role in relation classification.
Words closer to name entities have more influence on the determination of relation types.
We use position features, a combination of relative distances from current word to both entities, to identify entity pairs~\cite{zeng-etal-2014-relation}.
After concatenating position embeddings and word embeddings, the vector representation transforms a sentence into a matrix $S \in \mathbb{R}^{s\times d}$, where s is the sentence length and $d = d_{w} + d_{p}\times 2$.
For each word $w\in S = \begin{Bmatrix}
w_{1},w_{2},\cdots w_{n}
\end{Bmatrix}$, its embedding $\hat{w}_{i}$ is initialized as follows:
\begin{equation}
    \hat{w}_{i} = w_{i} \oplus p_{i1}\oplus p_{i2}
\end{equation}
where $w_{i}$ is the pre-trained word vector and $p_{i1}$, $p_{i2}$ are two corresponding position embeddings of the current word with two name entities. 
Symbol $\oplus$ indicates the concatenation operator.
The matrix $S$ is then fed into the CNN encoder.

\subsubsection{\textbf{CNN Encoder}}
Because convolutional neural networks can merge all local features and perform the prediction globally, we choose CNN to encode our input embeddings.
We learn the instance embedding as follows: 
\begin{equation}
    x_{i} = CNN(w_{i-\frac{n-1}{2}},\cdots ,w_{i+\frac{n-1}{2}})
\end{equation}
\begin{equation}
    \hat{x_{i}} = max(0, x_{i})
\end{equation}
\begin{equation}
    [s]_{j} = max\begin{Bmatrix}
[\hat{x_{1}}]_{j},\cdots, [\hat{x_{n}}]_{j}
\end{Bmatrix}
\end{equation}
where $CNN(\cdot)$ is a convolutional layer with window size $n$ over the word sequence.
A non-linear activation function ReLU is added after the convolutional layer.
Function max denotes max-pooling and $[\cdot]_{j}$ is the j-th value of a vector.

\begin{figure}[htp] 
 \center{\includegraphics[height=3.5cm,width=8cm]{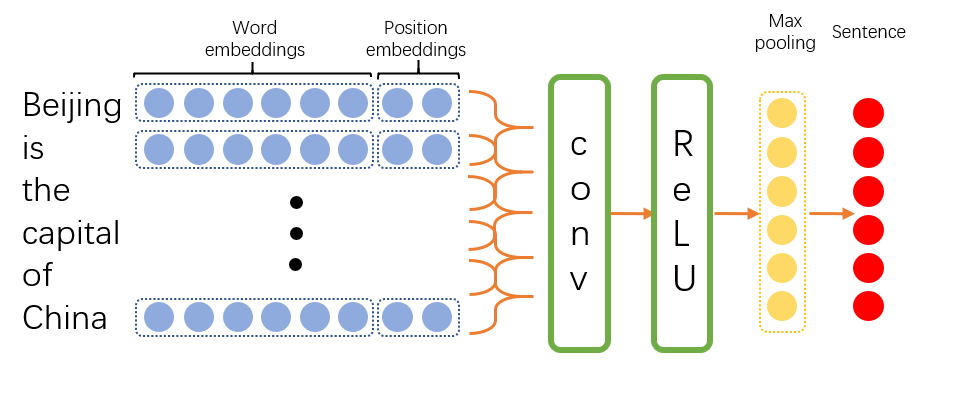}} 
 \caption{\label{fig:CNN} CNN Encoder} 
 \end{figure}

Figure~\ref{fig:CNN} shows the architecture of CNN encoder used in this paper.  
Due to time complexity, We simply use one convolutional layer, one non-linear layer, and one max pooling layer to get the sentence embedding.
The parameter settings are described in Section~\ref{sec: PS}

\subsubsection{\textbf{Side Information Embeddings}}
Label information and keywords in each sentence also play an essential role in improving the performance of relation classification.
For relations without any training sentences, hypernyms, labels and their corresponding synonyms are used as side information.
Side information embeddings are concatenated to the prototype for each relation after CNN encoder.
The final prototype including side information for each relation can be expressed as follows:
\begin{equation}
    {c_{i}}' = \left\{\begin{matrix}
r \oplus si_{h} \oplus si_{r} \oplus si_{k} \quad r\neq 0\\ 
si_{h} \oplus si_{r} \oplus si_{s} \qquad r = 0
\end{matrix}\right.
\end{equation}
where $r$ is the initial prototype for each relation, $si_{h}$ represents the side information from hypernyms, $si_{r}$ is the side information from relation types, $si_{k}$ is the side information from keywords in all training sentences of one relation type and $si_{s}$ is the synonyms for relation types.
Details for side information description and its extraction will be described in Section~\ref{section:SIA}.

\subsection{Side Information Extraction}\label{section:SIA}
Side information is the auxiliary information used to detect new relation types.
For relations with training sentences, side information is the hypernyms of two entities, relationship between two entities, and keywords from all training sentences with the same relation type.
For relations without training sentences, the side information is hypernyms of two entities by manually labeling, relation type itself, and synonyms of the relation types.
For query sentences, the side information is hypernyms of two entities and keywords extracted from the sentence.

In this section, we describe hypernyms extraction and keyword extraction in detail because the relationship can be easily obtained from labels, and synonyms of relation types can also be easily acquired through WordNet or other dictionaries~\cite{10.1093/ijl/3.4.235}.

\subsubsection{\textbf{Hypernyms Extraction}} \label{sec:ha}
\begin{figure}[htp] 
 \center{\includegraphics[height=1.4cm,width=8.5cm]{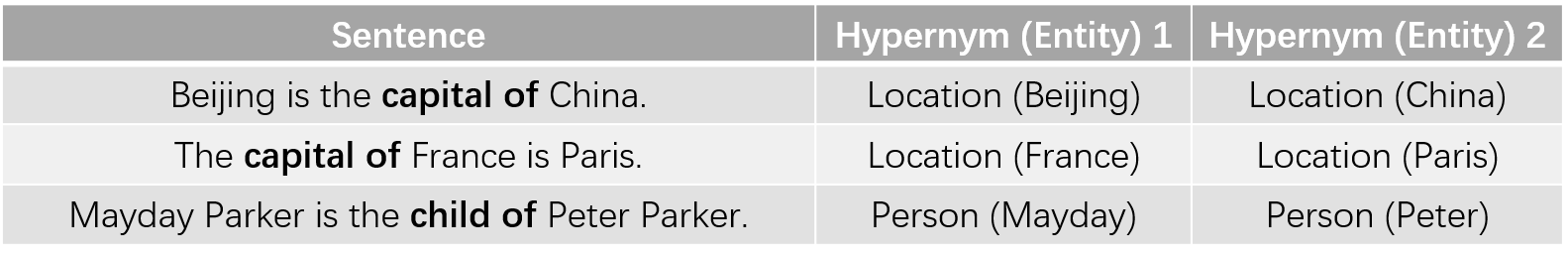}}
 \caption{\label{fig:Example} Example of sentences with different hypernyms.} 
 \end{figure}

A hypernym is the broad meaning of more specific words.
For example, an animal is a hypernym of a dog.
The hypernym of two entities in one sentence is extremely important for relation classification.
Figure~\ref{fig:Example} shows an example of different sentences with different hypernyms, indicating that hypernyms can help classify different relation types.
For example, relation \textit{capital\_of} can only occur between two locations, and relation \textit{child\_of} can only occur between two people.

Hypernyms of entities are not easy to acquire.
Some existing tools, such as WordNet can only acquire hypernyms from limited vocabularies.
In our experiments, less than 10\% of entities can achieve their corresponding hypernyms through WordNet.
Some previous works used entity types (hypernyms) defined by FIGER as side information~\cite{vashishth-etal-2018-reside, 10.5555/2900728.2900742}.
However, only 112 entity types are provided by FIGER, and only 38 of them are used as entity types by ~\cite{vashishth-etal-2018-reside}.
Most of the name entities from sentences in the real world can not get their hypernyms based on this list due to its fixed size and limited entity types.
Therefore, we provide an approach for extracting hypernyms through external help from the web.

Hypernyms can be discovered through the definition of entities.
We build an automatic hypernym extraction framework based on WordNet, Merriam Webster \footnote{https://www.merriam-webster.com/} and Wikidata \footnote{https://www.wikidata.org/}.
Merriam Webster includes a part of speech description to distinguish nouns of a person (biographical) from nouns of location (geographical).
In the real world, there are quite a number of relations that occur between these two hypernyms.
Wikidata provides definitions for different entities.
We crawl the definition for each entity through Wikidata and get the first Noun as hypernym.
For example, \textit{Jeff Bezos is the founder of Amazon.} Commerce is extracted as a hypernym for Amazon.
Most entities, including person, location or other nouns, can get their hypernyms through our proposed framework.
The entire framework of hypernym extraction is described in detail in Algorithm~\ref{alg:HA}.

\begin{algorithm}[!htbp]\label{Algorithm1}
\SetKwInOut{KIN}{Input}
\SetKwInOut{KOUT}{Output}
\caption{Hypernym Extraction}
\label{alg:HA} 
\KIN{sentences $\begin{Bmatrix}
x_{1},x_{2},x_{3},\cdots x_{n}
\end{Bmatrix}$ with same relation.}
\KOUT{hypernyms of two entities from one relation.}
Step 1: Initialize hypernyms to \textit{none}.\\
Step 2: Find hypernyms $\begin{Bmatrix}
h_{1}^{1}, h_{1}^{2},\cdots h_{1}^{n}
\end{Bmatrix}$ and $\begin{Bmatrix}
h_{2}^{1}, h_{2}^{2},\cdots h_{2}^{n}
\end{Bmatrix}$ of entities from WordNet.\\
Step 3: $h_{1} = major\begin{Bmatrix}
h_{1}^{1}, \cdots h_{1}^{n}
\end{Bmatrix}$, $h_{2} = major\begin{Bmatrix}
h_{2}^{1},\cdots h_{2}^{n}
\end{Bmatrix}$. \\
\eIf {$h == none$}{go to Step 4.}{End}
Step 4: Getting PoS descriptions $PD$ of entities $E = $ $\begin{Bmatrix}
e_{1}^{1}, \cdots e_{1}^{n}
\end{Bmatrix}$ and $\begin{Bmatrix}
e_{2}^{1}, \cdots e_{2}^{n}
\end{Bmatrix}$ from Merriam Webster. $h = Tokenize(PD)$\\
\eIf {$h == none$}{go to Step 5.}{End}
Step 5: Crawling definitions $D$ for $E$ from Wikidata. $h = $ first Noun of $Tokenize (D)$.\\
\end{algorithm}

\subsubsection{\textbf{Keywords Extraction}}
The keyword is another crucial factor of side information because it reflects the importance of the featured item.
TF-IDF (term frequency-inverse document frequency) is used for keyword extraction due to its efficiency~\cite{tfidf}.
It estimates the frequency of a word in one sentence over the maximum in a collection of sentences with the same relation type and assesses the importance of a word in one set of sentences.
For relations with training sentences, all sentences are aggregated as one document $d$, and TF-IDF is implemented based on the document.
Other models can also be used for keyword extraction.

\subsection{Prototypical Network with Side Information Embedding}\label{sec:wp}
Instead of adding a softmax layer directly after encoders for relation classification, we use prototypical networks to compute a prototype for each relation after encoders because some works show that prototypical networks work well for few-shot learning~\cite{NIPS2017_6996, article1}. 
They are simpler and more efficient than other meta-learning algorithms, making them suitable for few-shot or zero-shot learning tasks.
By comparing the distance between query sentences with prototypes for each relation, we can classify the relation.
In this section, we describe the prototypical network model and its transformation with weighted side information embedding for zero-shot learning to detect new relations.

The main idea for the prototypical network is to compute a prototype representing each relation.
Each prototype is the mean vector of embedded sentences belonging to one relation.
\begin{equation}
    c_{i}=\frac{1}{N}\sum_{i=1}^{N}f_{\phi }(x_{i})
\end{equation}
where $c_{i}$ represents the prototype for each relation $r_{i}$ and $f_{\phi }$ is an embedding function, which is a CNN encoder in our model.
Instead of concatenating all hypernyms and keywords directly after each prototype, we argue that not all keywords are of equal importance.
To determine a more accurate representation for each relation, we calculate a weighted side information embedding for each relation.
The equation of side information embedding $si$ is as follows:
\begin{equation}
   si = f(\frac{h_{1}+h_{2}}{2})\oplus f(k_{1})\oplus \cdots \oplus f(k_{n})\oplus K
\end{equation}
\begin{equation}
    K = \sum_{m-n}^{m}(\frac{\alpha _{i}}{\sum_{i=m-n}^{m}\alpha _{i}}f(k_{i}))
\end{equation}
where $f(\cdot)$ is a word embedding model, $h_1$ and $h_2$ are two hypernyms for name entities and $k_i$ denotes the keyword.
Symbol $\oplus$ is the concatenation operator, n is determined by exploration search, $m$ is the total number of keywords and $\alpha_i$ is a calculated weight by:
\begin{equation}
    \alpha_{i} = \frac{count(k,s)}{size(s)}\cdot log(\frac{N}{sentence(k, S)})
\end{equation}
where $s$ is each instance and $N$ is the number of instances in a relation.
The final representation for each prototype with side information embedding $ps_{i}$ can be expressed by:
\begin{equation}
    ps_{i} = c_{i} \oplus si_{i}
\end{equation}
The probabilities of the relations in $\Re$ for a query instance $x$ is computed as follows:
\begin{equation}
    p_{\phi }(y=ps_{i}|x)=\frac{exp(-d(f_{\phi }(x),ps_{i}))}{\sum_{ps_{i}^{'}}exp(-d(f_{\phi }(x),ps_{i}^{'}))}
\end{equation}
where $d(.)$ is Euclidean distance function as below:
\begin{equation}
    d(f_{\phi }(x),ps_{i})=\sqrt{\sum_{i=1}^{n}(ps_{i}-f_{\phi }(x))^{2}}
\end{equation}

We use Euclidean distance instead of cosine similarity for distance calculation because previous work shows that Euclidean distance can improve performance substantially over cosine similarity~\cite{NIPS2017_6996}.
We have not added any attention layer in our final model because (1) previous work shows there is little improvement on performance compared with vanilla prototypical networks~\cite{article1}; (2) Ablation study in Section~\ref{sec:fewrel} shows there is no improvement on ZSLRC with attention layers.

For the zero-shot learning task, each relation is given the embedding for side information of the relation rather than a small number of labeled training sentences.
We take the embedding of side information into a shared space to serve as the prototype for each relation.
The core idea in traditional prototypical networks is to use an average embedding to represent a class~\cite{NIPS2017_6996, article1}. 
If there are no training data in that class, a high-level description of the class is used to represent that class. We modify prototypical networks to deal with both relations with training sentences and relations without training sentences.
The difference between traditional prototypical networks and our proposed model is that they calculate the distance between the query sentence and prototype of each class to find the nearest one.
Our proposed model first decides the query sentence is in a class with training data or the one without any training data based on a threshold.
The reason is that finding the nearest distance directly based on all classes (with training data and without training data) is not fair for the class without training data because the high-level description is too general that it always has a longer distance compared with the classes which have training data.

We modify the prototypical network as follows:
We first compare the distance between an input sentence with each prototype of known relations.
The key mechanism for extracting new relations is that if the above distance is larger than a threshold, we consider the sentence has a new relation.
Then we take the side information embedding of the input sentence and compare the distance between it with prototypes for new relations.
Then we use a softmax layer to compute the probabilities for each new relation.
The threshold selection is essential because it influences the decision of a relation type as an existing relation or a new relation.
We implement a grid search to select the optimal threshold on the validation set.
The entire framework of ZSLRC model to deal with a combination of known relations and new relations is described in Algorithm~\ref{alg:model}. 

\begin{algorithm}[!htbp]
\SetKwInOut{KIN}{Input}
\SetKwInOut{KOUT}{Output}
\caption{Algorithms for New Relation Extraction}
\label{alg:model} 
\KIN{prototype for each relation $c_{i}$, testing sentence $x$, threshold $t$.}
\KOUT{relationship $r$ of $x$.}
Distance Calculation. $d(f_{\phi }(x),c_{i})$.\\
Take $v = exp(-d(f_{\phi }(x),c_{i}))$.\\
\eIf {$v > t$}{Classification of known relations. $r = argmax(\frac{v}{\sum _{c_{i}^{'}}v^{'}})$}
{Take side information embedding. $f_{\phi }(x)[SI\_DIM:]$\\
Distance Calculation. $d(f_{\phi }(x)[SI\_DIM:],c_{i})$.\\
Take $v_{new} = exp(-d(f_{\phi }(x)[SI\_DIM:],c_{i}))$.\\
Softmax of new relations. $r_{new} = argmax(\frac{v_{new}}{\sum _{c_{i}^{'}}v_{new}^{'}})$}
\end{algorithm}

\section{Experiments} \label{Experiment}
In this section, we conduct several experiments on two public datasets: NYT~\cite{10.1007/978-3-642-15939-8_10} and FewRel~\cite{han-etal-2018-fewrel} to show that our proposed model outperforms other existing models on both a noisy dataset with a large number of training sentences and a clean dataset with few training sentences.
We design experiments for generalized zero-shot learning tasks and provide a detailed analysis to show the effectiveness and advantages of our proposed model.

\subsection{Datasets and Evaluation Metrics}
In our experiments, we evaluate our model over two widely used datasets: the NYT dataset~\cite{10.1007/978-3-642-15939-8_10} and FewRel~\cite{han-etal-2018-fewrel} dataset.
In the following, we describe each dataset in detail.

\begin{itemize}
	\item \textbf{\textit{NYT~\cite{10.1007/978-3-642-15939-8_10}.}}
The NYT dataset was generated by aligning Freebase relations with the New York Times corpus (NYT).
There are 53 possible relationships in total.
It is an unbalanced noisy dataset because all the relationships have a different number of sentences.
	\item \textbf{\textit{FewRel~\cite{han-etal-2018-fewrel}.}}
The FewRel dataset is a human-annotated few-shot RC dataset consisting of 80 types of relations, each of which has 700 instances.
\end{itemize}

To fairly compare the performance of our proposed model with other state-of-the-art models in supervised learning and few-shot learning tasks, we use the same training, validation and testing set of NYT dataset and same training and validation set of FewRel.
We evaluate our proposed model on the validation set of FewRel because the test set is not available directly.
In order to properly evaluate the performance of our proposed model in a zero-shot learning task, we re-split the above two public datasets for training, validation and testing set.
Details of dataset re-splitting and experiment design are introduced in section~\ref{ED}.
Note that we do not use any other clean, supervised dataset such as SemEval-2010 Task 8 (SemEval) because this dataset only contains 19 kinds of relations, which is less persuasive when re-splitting the dataset to evaluate the performance of our proposed model in zero-shot learning task~\cite{hendrickx-etal-2010-semeval}.

The evaluation metrics adopted in this paper are the standard micro Accuracy (Acc.), Precision (Prec.), Recall (Rec.) and F1-score, similar to those used for the baseline.

\subsection{Experiment Design} \label{ED}
In a real-world scenario, there exist both kinds of relations with training instances and without any training instances.
To make it simple and clear to understand, we call the relations with training instances known relations and the relations without any training instances new relations in the following discussion.
To evaluate the effectiveness and robustness of our proposed model in a zero-shot learning task, we design the testing cases to contain different percentages (from 0\% to 100\% with a step of 10\%) of new relations.
Note that 0\% means a thoroughly supervised learning or few-shot learning scenario, whereas 100\% means a completely zero-shot learning scenario.
The experiment design for zero-shot learning relation classification follows the criteria of zero-shot text classification; the different rates of unseen classes are used in testing cases~\cite{zhangkumjornZeroShot}.

\noindent\textbf{\textit{NYT~\cite{10.1007/978-3-642-15939-8_10}.}}
NYT is an unbalanced noisy dataset with 53 different relationships in total.
We added initial training, validation and testing sets together and re-split the dataset into ten types of relations for the training pool. Each relation has over 10k sentences, and the rest relations are for the validation pool and testing pool.
In the training pool, we take 10k sentences of each relationship for training, and the rest types of relations are used to validate and test known relations.
In all, we have 100k sentences of 10 relationships in total for training, 13k sentences of known relations, and 5k sentences of new relations for validation and testing.
For example, if a new relation \textit{capital\_of} is allocated to testing set, no \textit{capital\_of} sentences appear in training set.

\noindent\textbf{\textit{FewRel~\cite{han-etal-2018-fewrel}.}}
FewRel dataset has 80 types of relations with 700 instances each.
We re-split the dataset into 40 types of relations for training and 40 types of relations for testing.
There are no overlapping relations among the training and testing sets.
To evaluate our proposed model in a real-world scenario (a combination of known and new relations in the testing set), we take 300 instances from each relation type in the training set to make a testing pool containing known relations.
In total, we have 40 relations, and each relation has 400 instances in the training pool, 40 known relations. Each relation has 300 instances in the testing pool, 40 new relations, and each relation has 700 instances in the testing pool.

\subsection{Parameter Settings} \label{sec: PS}
For all the models, we use the pre-trained word embeddings with a 50-dimensional Glove model (6B tokens, 400K vocabulary) and a randomly initialized 5-dimensional position embedding on NYT corpus for initialization~\cite{pennington2014glove}.
Both word embeddings and position embeddings are trainable during training.
The number of feature maps in the convolutional layer is 800, and the side information embedding dimension is 300.
We experimentally study the effects of two crucial parameters on our model, learning rate $\alpha$ and threshold $t$.
We use a grid search to select the optimal learning rate $\alpha$ for SGD among $\begin{Bmatrix}
1e-1, 1e-2, 1e-3, 1e-4
\end{Bmatrix}$ for minimizing the loss, the threshold $t$ for determining a new relation among $\begin{Bmatrix}
2e-08, 7e-08, 2e-07, 7e-07
\end{Bmatrix}$ on a validation set with 20\% of new relations.
The range of threshold is determined by the minimum and maximum values of $e^{-d}$ on a validation set, where $d$ is the Euclidean distance between query sentence and prototype for each relation.
For other parameters, we follow the settings used in previous works so that our model can be fairly compared with these models~\cite{zeng-etal-2014-relation, article1}.
Table~\ref{tab:parameter} shows parameters used in our experiment.

\begin{table}[]
\caption{Parameter Settings}
\label{tab:parameter}
\begin{tabular}{lr}
\hline
Parameter                            & Value     \\ \hline
Word Embedding Dimension $d_{w}$     & 50        \\
Position Embedding Dimension $d_{p}$ & 5         \\
Side Information Embedding Dimension $d_{si}$  & 300\\
Hidden Layer Dimension $d_{h}$       & 800       \\
Convolutional Window Size $n$        & 3         \\
Batch Size                           & 1         \\
Initial Learning Rate $\alpha$       & 0.01      \\
Weight Decay                         & $10^{-5}$ \\
Threshold $t$                        & 2e-08    \\ \hline
\end{tabular}
\end{table}

\subsection{Results}
\subsubsection{\textbf{Baseline Methods}}
We compare our proposed model to several state-of-the-arts models in both supervised learning and few-shot learning tasks. For a supervised learning task on the NYT dataset, we compare our model with \textbf{CDNN}, which first proposed the idea of position embedding~\cite{zeng-etal-2014-relation}.
The reason we choose this model to make the comparison is that we both use similar CNN encoders so that the improved performance of our model is not because of using any better encoders such as BERT~\cite{Devlin2019BERTPO}.
The reported result for \textbf{CDNN} is our re-implementation on NYT because the source code is not available, and their original report is the evaluation on other datasets~\cite{zeng-etal-2014-relation}.
The reported result for the \textbf{REDN} is from the original published literature ~\cite{Li2020DownstreamMD}.
Note that \textbf{REDN} is a relation classification model using the given name entities, and we only copy the result of the single relation classification of this paper so that we could make a fair comparison.
For few-shot learning task on FewRel dataset, we compare our model with \textbf{Meta Network}, \textbf{GNN}, \textbf{SNAIL}, \textbf{Proto}, \textbf{Proto-HATT} and \textbf{Proto-CATT(CNN)}.
The six baselines above on the FewRel dataset are reported by~\cite{few-shot2020}, which are all current state-of-the-art FSL models.
Note that the above FSL model Proto-HATT and our proposed model use the same pre-trained word embedding model 50-dimension GloVe, CNN encoders and same training parameters only except batch size and hidden layer dimension.
For zero-shot learning, we compare our proposed model with the re-implemented \textbf{CDNN}, \textbf{REDN}, \textbf{Proto} and \textbf{Proto-HATT} on our re-splitted NYT and FewRel datasets to show the effectiveness and robustness of our proposed model.

\subsubsection{\textbf{Results on NYT}}\label{sec:nyt}
Table~\ref{tab:NYT} demonstrates that our proposed model achieves a substantial gain in precision, recall and F1-score over other baselines for the supervised learning task.
We compare ZSLRC model with CDNN~\cite{zeng-etal-2014-relation} as both models use a CNN encoder. The results show that ZSLRC achieves a significant performance improvement on precision, recall and F1-score.
Our proposed ZSLRC also outperforms a recently proposed method (REDN)~\cite{Li2020DownstreamMD} by \textbf{3\%} precision, \textbf{3.9\%} recall and \textbf{3\%} F1-score though REDN uses BERT encoder. This is important to note because BERT-based sentence encoders have significantly outperformed other sentence encoders including our proposed one-layer CNN based type~\cite{few-shot2020}.

\begin{figure}[htp] 
 \center{\includegraphics[height=6cm,width=7cm]{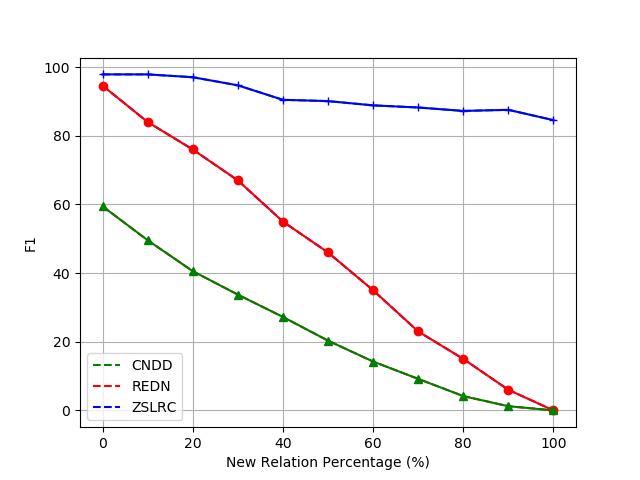}}
 \caption{\label{fig:supervised} F1-score of ZSLRC when different proportions of new relations appear in NYT dataset.} 
\end{figure}

The achieved performance improvement indicates that the proposed side information is competitively beneficial for relation classification.
To evaluate our proposed model in a real-world scenario, we re-split the NYT dataset and use 40+ relations as new relations with no labeled training data.
As is shown in Figure~\ref{fig:supervised}, 0\% of new relations means it is a supervised learning task and all relations in the testing set have corresponding labeled training data.
100\% of new relations means it is a conventional zero-shot learning task, and all relations in the testing set do not have any labeled training data.
We compare the performance of our proposed model with CDNN~\cite{zeng-etal-2014-relation} and REDN~\cite{Li2020DownstreamMD} as we vary the percentage of new relations in the testing set.
As shown in Figure~\ref{fig:supervised}, the F1-score of both CDNN and REDN decrease when the percentage of new relations increase. This is because the model can not detect new relations and instead classifies the new relation as one of the existing relations in the training set.
That is why the F1-score becomes zero when the new relation percentage is 100\%.
The F1-score of our proposed model ZSLRC only drops around 15\% from a fully supervised case to a zero-shot case, indicating that our model is effective and sufficiently robust when dealing with new relations.

\begin{table}[]
\caption{Results of different models on NYT (\%).  Our re-implementation is marked by $*$.}
\label{tab:NYT}
\begin{tabular}{lccc}
\hline
Model & Precision     & Recall        & F1            \\ \hline
CDNN$^{*}$ \cite{zeng-etal-2014-relation}  &  46.4             &  52.7             &    45.8           \\
REDN \cite{Li2020DownstreamMD}  & 95.1          & 94.0          & 94.6          \\ \hline
ZSLRC & \textbf{98.1} & \textbf{97.9} & \textbf{97.6} \\\hline
\end{tabular}
\end{table}

\begin{table}[]
\caption{Ablation Results on NYT dataset (Accuracy\%)}
\label{tab:Ablation}
\begin{tabular}{lccccc}
\hline
     & 10\%  & 30\%  & 50\%  & 70\%  & 90\%  \\ \hline
ZSLRC(HE)   & 88.94 & 70.57 & 52.12 & 33.87 & 15.48 \\
ZSLRC(KE)   & 93.12 & 82.22 & 71.00 & 60.47 & 49.07 \\
ZSLRC(SIE)  & 93.86 & 85.14 & 81.91 & 78.79 & 72.57  \\
ZSLRC(WSIE) & \textbf{96.64} & \textbf{94.46} & \textbf{92.14} & \textbf{91.82} & \textbf{89.3}  \\ \hline
\end{tabular}
\end{table}

\begin{figure}[htp] 
 \center{\includegraphics[height=6cm,width=7cm]{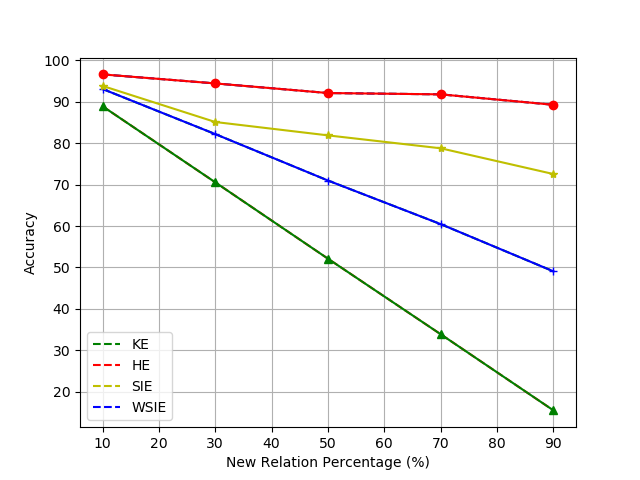}}
 \caption{\label{fig:ablation} Ablation study of ZSLRC on NYT dataset.} 
\end{figure}

To investigate the contribution of different side information embeddings in ZSLRC, we conduct an ablation study in zero-shot learning settings by adding each component, including hypernyms embedding (HE), keywords embedding(KE), side information embedding(SIE) and weighted side information embedding(WSIE).
Table~\ref{tab:Ablation} shows the results of the ablation study different proportions of new relations in the testing set. 
We find out that all kinds of side information embedding help detect new relations.
Only adding hypernyms embedding to the model can help detect new relation classes.
However, the accuracy rate drops significantly from 88.94\% in 10\% of new relations to 15.48\% in 90\% of new relations.
This is because hypernyms only represent the main categories for name entities and could help classify the relations roughly without training instances.
Compared with hypernyms embedding, keywords embedding achieves much better performance because keywords (keywords extracted from training instances of seen class and synonyms of labels of unseen class) represent discriminative features of each instance, shorten the distance between query instance and prototype.
Nevertheless, the performance of ZSLRC(KE) still drops considerably when the percentage of new relations increase.
ZSLRC(SIE) achieves a significant accuracy performance improvement.
Side information embedding is a combination of hypernyms embedding and keywords embedding.
It represents high-level information of the instance, shortening the distance of instances with the same relation.
As shown in Table~\ref{tab:Ablation}, it is more robust when the percentage of the new relation class increases.
Since we assume that not all side information is of equal importance, we also implement ZSLRC with weighted side information added to the model as introduced in Section~\ref{sec:wp}.
This model achieves the best performance.
Besides the high accuracy performance with any proportions of new relations, it is also robust enough that it only drops 7.3\% accuracy rate from 10\% of new relations to 90\% of new relations.
Figure~\ref{fig:ablation} indicates the accuracy improvement and robustness of weighted side information embedding.
When the proportions of new relation increase, accuracy of ZSLRC with weighted side information embedding drops less than the other models.

\subsubsection{\textbf{Results on FewRel}}\label{sec:fewrel}

\begin{table*}[!htb]
\caption{Results of Accuracy Comparison Among Models (\%)}
\label{tab:fsl}
\begin{threeparttable}
\begin{tabular}{lcccccc}
\hline
Model             & 5 way 1 shot   & 5 way 5 shot   & 5 way 10 shot  & 10 way 1 shot  & 10 way 5 shot  & 10 way 10 shot \\ \hline
Meta Network$^*$      & 64.46 $\pm$ 0.54          & 80.57 $\pm$ 0.48         & -              & 53.96  $\pm$ 0.56        & 69.23   $\pm$ 0.52       & -              \\
GNN$^*$               & 66.23   $\pm$ 0.75       & 81.28  $\pm$   0.62     & -              & 46.27   $\pm$ 0.80       & 64.02   $\pm$ 0.77       & -              \\
SNAIL$^*$             & 67.29 $\pm$ 0.26         & 79.40   $\pm$ 0.22        & -              & 53.28 $\pm$ 0.27         & 68.33    $\pm$ 0.25      & -              \\
Proto(CNN)             & 73.62   $\pm$ 0.20       & 85.78   $\pm$ 0.16       & 88.45   $\pm$ 0.10       & 60.96   $\pm$ 0.22       & 75.38  $\pm$ 0.19        & 78.71     $\pm$ 0.11     \\
Proto-HATT(CNN)         & 74.68   $\pm$ 0.18       & 86.73  $\pm$ 0.12        & 89.64     $\pm$ 0.12     & 61.61    $\pm$ 0.16      & 77.04   $\pm$  0.12     & 79.99  $\pm$ 0.11        \\
Proto-CATT(CNN)          & -              & 87.48  $\pm$ 0.12        & 89.28   $\pm$0.08       & -              & 77.46  $\pm$ 0.13        & 80.39     $\pm$ 0.14     \\ \hline
\textbf{ZSLRC(CNN)} & \textbf{75.83$\pm$0.17} & \textbf{87.84$\pm$0.12} & \textbf{89.67$\pm$0.12} & \textbf{63.54$\pm$0.14} & \textbf{77.64$\pm$0.11} & \textbf{80.69$\pm$0.10} \\ \hline
\end{tabular}
\begin{tablenotes}

      \item Note that to fairly compare the performance of each model, we only compare the models with the same 50-dimension GloVe embedding and CNN encoders of the same parameters. Better results can be achieved through the BERT encoder.
    \end{tablenotes}
    \end{threeparttable}
\end{table*}

\begin{table*}[]
\caption{Ablation Results on FewRel dataset (\%).}
\label{tab:ablation}
\begin{tabular}{lcccccc}
\hline
Model                & 5 way 1 shot   & 5 way 5 shot   & 5 way 10 shot  & 10 way 1 shot  & 10 way 5 shot  & 10 way 10 shot \\ \hline
Proto(CNN)           & 73.62     $\pm$ 0.20      & 85.57     $\pm$ 0.14     & 88.17   $\pm$ 0.10       & 62.22    $\pm$ 0.32      & 75.01    $\pm$ 0.16      & 78.50    $\pm$ 0.11      \\
ZSLRC(HE)            & 75.66   $\pm$ 0.14       & 86.55    $\pm$ 0.13      & 88.98   $\pm$ 0.10       & 63.28 $\pm$ 0.20         & 76.58   $\pm$ 0.06       & 79.93   $\pm$ 0.05       \\
ZSLRC(KE)            & 74.57   $\pm$ 0.08      & 86.70   $\pm$ 0.17       & 89.09  $\pm$ 0.11        & 62.39   $\pm$ 0.12       & 76.99    $\pm$0.20      & 80.06     $\pm$ 0.09     \\
ZSLRC(SIE)           & 75.56    $\pm$ 0.12      & 87.34     $\pm$ 0.14     & 89.17   $\pm$ 0.13       & 63.02         $\pm$ 0.15 & 77.16     $\pm$ 0.12     & 80.34   $\pm$ 0.10       \\
\textbf{ZSLRC(WSIE)} & \textbf{75.83$\pm$0.17} & \textbf{87.84$\pm$0.12} & \textbf{89.67$\pm$0.12} & \textbf{63.54$\pm$0.14} & \textbf{77.64$\pm$0.11} & \textbf{80.69$\pm$0.10} \\
ZSLRC(WSIEA)         & 75.58    $\pm$ 0.15      & 87.16     $\pm$ 0.16     & 89.17  $\pm$ 0.15        & 62.85   $\pm$ 0.18       & 76.71   $\pm$ 0.14       & 80.18   $\pm$0.11       \\ \hline
\end{tabular}

\end{table*}

\begin{figure}[htp] 
 \center{\includegraphics[height=6cm,width=7cm]{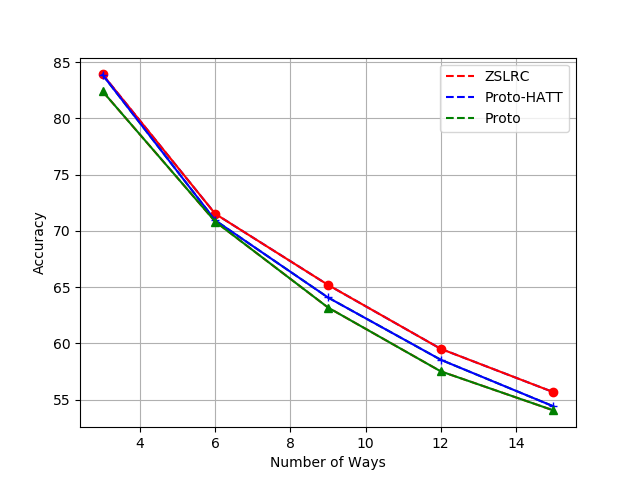}}
 \caption{\label{fig:fsl} Accuracy of our proposed model in different N-way One-shot tasks.} 
 \end{figure}

The evaluation results of few-shot learning on FewRel are shown in Table~\ref{tab:fsl}.
Note that results with $^*$ are reported in ~\cite{han-etal-2018-fewrel}.
The result of Proto-CATT model is copied from their original paper because of no public code ~\cite{few-shot2020}.
We re-implement Proto and Proto-HATT with all parameters the same except hidden layer dimension.
Both Proto-HATT and Proto-CATT are using CNN encoders and attention layers to help improve the performance.
To fairly compare the effectiveness of side information embedding, we only compare our models with other state-of-the-art models using CNN encoders with attention layers. 
Each task is provided with a set of k labeled sentences from each of N classes that have not previously been trained upon.
We conduct the experiments of N-way K-shot few-shot learning tasks following the method introduced in~\cite{Nichol2018OnFM}.
Table~\ref{tab:fsl} shows that ZSLRC (without any attention layer) outperforms the other state-of-the-art models using multiple attention layers on several N-way K-shot tasks, especially for 1-shot cases.
The accuracy of our proposed model on 5-way 1-shot and 10-way 1-shot tasks are 75.83\% and 63.54\%, which is 1.15\% higher and 1.93\% higher than the model Proto-HATT.
Next, we investigate ZSLRC performance on N-way one-shot learning. Figure~\ref{fig:fsl} demonstrates changes in accuracy as the number of ways changes in comparison with two state-of-the-art models. 
As the number of classes increases, the accuracy drops, but our proposed model has a slower dropping rate than other models.
We conjecture that both the increased difficulty of a larger number of ways and the side information embedding we have proposed enables the ZSLRC to make more fine-grained decisions and is therefore more robust to the increased complexity introduced by more classes.

\begin{figure}[htp] 
 \center{\includegraphics[height=6cm,width=7cm]{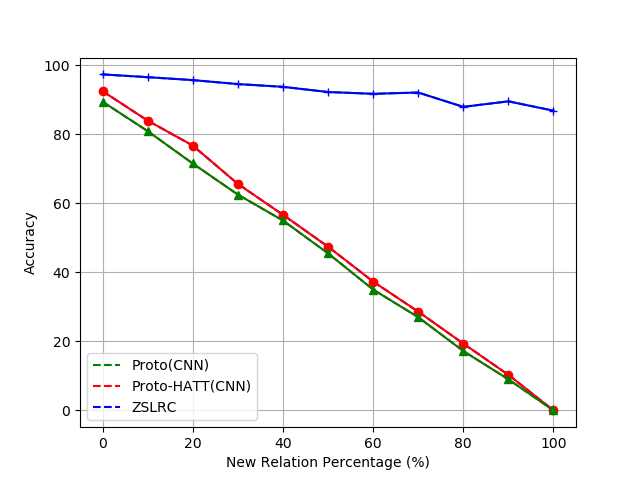}} 
 \caption{\label{fig:zsl0} Accuracy of ZSLRC when different proportions of new relations appear in re-splitted FewRel dataset.} 
 \end{figure}

To evaluate the effectiveness and robustness of ZSLRC in a generalized zero-shot learning task, we evaluate our models on the re-splitted FewRel dataset.
To test the effectiveness and robustness of our proposed model, we compare our proposed model ZSLRC with Proto(CNN) and Proto-HATT(CNN) ~\cite{article1} in zero-shot settings described in Section~\ref{ED}.
Figure~\ref{fig:zsl0} shows the performance of ZSLRC in a real world scenario with different percentages of new relations on re-splitted FewRel dataset. 
The accuracy of ZSLRC only drops from 97.3\% to 86.8\%, indicating the effectiveness and robustness of our proposed model for recognizing new relations in the real world.
We show that zero-shot learning to new relation types is possible and we set the bar for future work on this task.

We also conduct an ablation study on FewRel dataset to learn the effectiveness of weighted side information embedding.
Besides the models introduced in Section~\ref{sec:nyt}, we also implement a new model with attention layers for weighted distance (WSIEA), to investigate the influence of the attention layer.
Table~\ref{tab:ablation} shows the results of ablation study.
We can observe that all kinds of side information embedding contribute to the performance of ZSLRC.
There is a big accuracy performance improvement when hypernyms embedding introduced in Section~\ref{section:SIA} is added to the model because hypernyms represent a general embedding for different name entities, which will decrease the variance from different word embeddings, leading to a shorter distance.
Keyword embedding also contributes significantly to the performance, indicating the importance of keywords to side information embedding.
Similar to the ablation result on NYT dataset as shown in Section~\ref{sec:nyt}, using side information embedding helps improve the performance and the model with weighted side information embedding achieves the best performance.
We also added an attention layer built by three neural network layers and a softmax layer on top of each prototype to calculate linear separability based on the distribution of each prototype's sentence representations.
However, there is no improvement of the attention layer.
We conjecture that the weighted side information embedding has already captured each relation's vital feature.
In this way, merely using side information embedding helps simplify the model's architecture, reducing the complexity of several neural network layers by attention mechanism.

\section{Conclusion and Future Work} \label{Con}
We propose ZSLRC\footnote{Implementation details can be accessed via: https://github.com/gjiaying/ZSLRC}, a zero-shot learning relation classification framework based on modified prototypical networks. ZSLRC can detect new relations with no corresponding labeled data available for training.
ZSLRC utilizes weighted side information constructed from labels, keywords and hypernyms of entities extracted from our proposed automatic hypernym extraction framework.
We evaluate our model on supervised learning, few-shot learning and zero-shot learning tasks. The results demonstrate that our proposed ZSLRC outperforms other state-of-the-art models in all tasks. 
In addition, the results demonstrate the effectiveness and robustness of our proposed model. 
In future work, we plan to explore the following directions:
(1) Due to the surprising performance improvement contributed by side information embedding, we will explore different ways to embed side information, leading to learning different representations of each prototype (relation). 
(2) We will explore using other popular sentence encoders such as BERT to improve the performance for relation classification.

\bibliographystyle{ACM-Reference-Format}
\bibliography{sample-base}

\appendix

\end{document}